\def\year{2021}\relax
\documentclass[letterpaper]{article} 
\usepackage{aaai21}  
\usepackage{times}  
\usepackage{helvet} 
\usepackage{courier}  
\usepackage[hyphens]{url}  
\usepackage{graphicx} 
\usepackage{natbib}  
\usepackage{caption} 
\usepackage{amsfonts,amssymb}
\usepackage{booktabs} 
\usepackage{url}

\usepackage{algorithm}
\usepackage{algorithmicx}
\usepackage{algpseudocode}
\usepackage{amsmath}
\usepackage{wrapfig}
\usepackage{graphicx}
\usepackage{caption}
\usepackage{multirow}
\usepackage{color}
\usepackage[switch]{lineno}  %

\newcommand{\xw}[1]{\textcolor{blue}{#1}}

\title{Overcoming Catastrophic Forgetting in Graph Neural Networks}

\author {
        Huihui Liu, \,\, 
        Yiding Yang, \,\, 
        Xinchao Wang\footnote{Corresponding Author.}\\ 
}
\affiliations {
    Stevens Institute of Technology\\
    \{hliu79, yyang99, xinchao.wang\}@stevens.edu
}

\begin{document}

\maketitle

\begin{abstract}
Catastrophic forgetting refers to the tendency that 
a neural network ``forgets'' the previous learned 
knowledge upon learning new tasks.
Prior methods have been focused on overcoming this problem 
on convolutional neural networks (CNNs), where the input samples 
like images lie in a grid domain, but have largely overlooked 
graph neural networks (GNNs) that handle non-grid data.
In this paper, we propose a novel scheme 
dedicated to overcoming catastrophic forgetting problem and 
hence strengthen continual learning in GNNs.
At the heart of our approach is a generic module,
termed as topology-aware weight preserving~(TWP),
applicable to arbitrary form of GNNs in a plug-and-play fashion.
Unlike the main stream of CNN-based continual learning methods
that rely on solely slowing down the updates of
parameters important to the downstream task,
TWP explicitly explores the local structures of the input graph,
and attempts to stabilize the parameters playing 
pivotal roles in the topological aggregation. 
We evaluate TWP on different GNN backbones over several datasets, 
and demonstrate that it yields performances
superior to the state of the art.
Code is publicly available at~\url{https://github.com/hhliu79/TWP}. 


\end{abstract}

\section{Introduction}
Deep neural networks have demonstrated unprecedentedly gratifying 
results in many artificial intelligence tasks. 
Despite the encouraging progress,
in the \emph{continual learning} scenario
where a model is expected to learn
from a sequence of tasks and data,
deep models are prone to the 
\emph{catastrophic forgetting} problem~\cite{mccloskey1989catastrophic,ratcliff1990connectionist,mcclelland1995there,french1999catastrophic}.
For instance, when a model is first trained to convergence on one task and then trained on a second task, 
it ``forgets'' how to perform the prior task and consequently leads
to inferior results.

\begin{figure}[t]
\setlength{\abovecaptionskip}{0.05cm}
\centering
\includegraphics[width=0.45\textwidth]{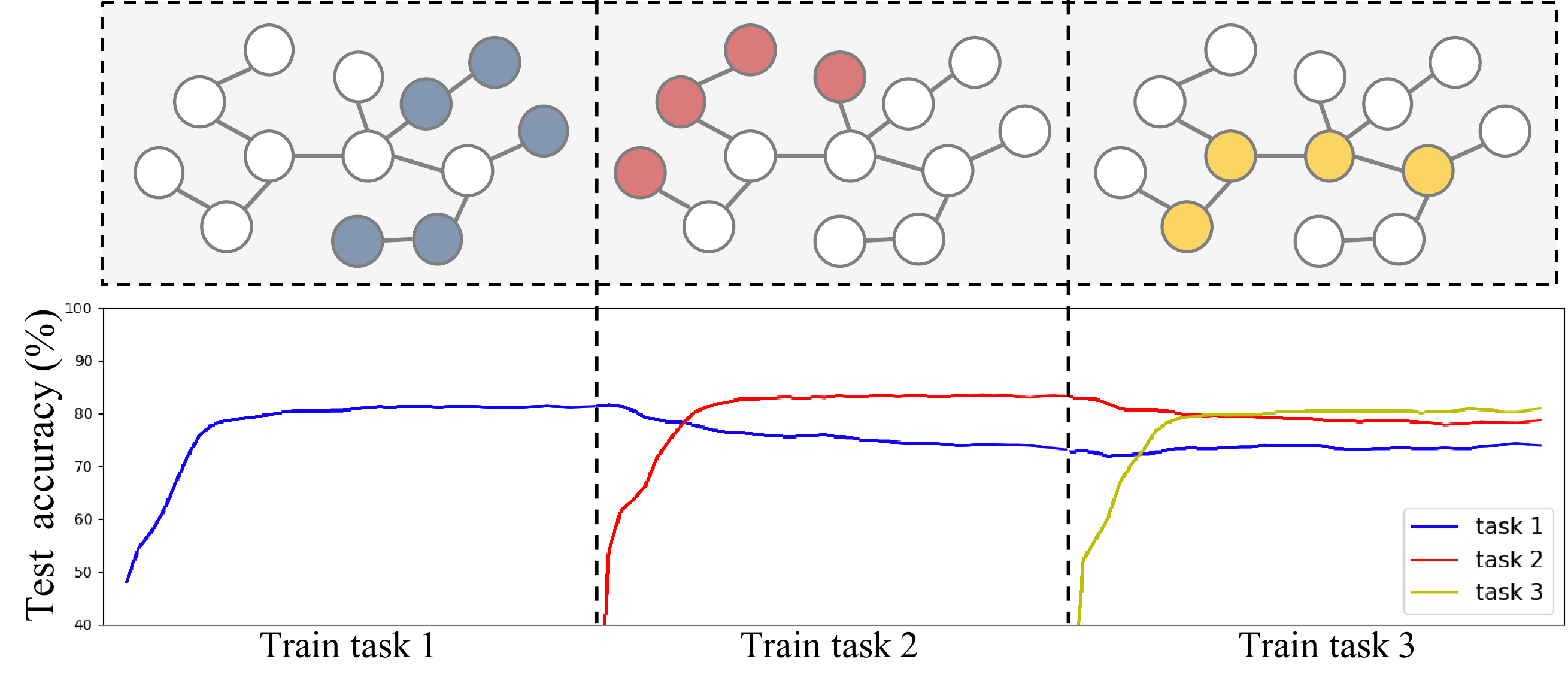}
\caption{Illustration of catastrophic forgetting 
on graph attention networks (GATs) models on the Corafull dataset based on a CNN-based method, EWC~\cite{kirkpatrick2017overcoming}. 
Top: Blue, red, and yellow nodes are used 
for task 1, 2, and 3, respectively.
Bottom: The test performance of each task 
when learning new tasks gradually.
}
\vspace{-1.7em}
\label{cf}
\end{figure}

Many recent endeavours have been made towards alleviating 
catastrophic forgetting
or preserving the performance
on the older task upon learning newer ones. 
Earlier approaches rely on a rehearsal scheme,
in which a part of samples from previous tasks
are stored and replayed in learning the new task~\cite{lopez2017gradient,rebuffi2017icarl}.
Regularization-based methods, as another popular line of work,
precludes the key parameters of the previous tasks
from undergoing dramatic 
changes~\cite{kirkpatrick2017overcoming, li2017learning, aljundi2018memory}.
Parameter-isolation methods, on the other hand, 
focus on assigning different subsets of parameters
for different tasks
and reusing knowledge from previous tasks~\cite{rusu2016progressive, fernando2017pathnet, yoon2017lifelong}.

Existing deep-learning-based continual learning methods,
in spite of their impressive results,
have been focusing on convolutional neural networks (CNNs) 
with the grid data like images as input.
However, many real-world data 
lie in the non-grid domain and
take the form of graphs.
Examples include but are not limited to 
such as citation networks, social networks~\cite{newman2004finding},
and biological reaction networks~\cite{pavlopoulos2011using}.
To handle data in the irregular domain, 
various graph neural networks (GNNs) have been 
proposed~\cite{kipf2016semi,hamilton2017inductive} 
to explicitly account for the topological interactions between 
nodes in the graph.

Yet unfortunately, GNNs suffer from catastrophic forgetting as well.
As demonstrated in Figure~\ref{cf}, 
the state-of-the-art graph attention networks (GATs),
when applied to a sequence of three node-classification tasks,
significantly deteriorate 
as new tasks are learned.
A straightforward solution towards handling this issue 
is to directly apply  continual learning methods designed for CNNs 
on the graph data, by 
{treating each node as a single sample like an image for CNNs.}
In doing so, however, the CNN-based methods merely focus
on individual nodes of the graph but neglect the topological structure
and the interconnections between nodes,
which play pivotal roles in  propagating and aggregating information
in GNNs. In the case of Figure~\ref{cf}, 
this approach leads to a performance drop
of 7.48$\%$ on the accuracy of task~1
upon learning task~3. 
In this paper, we propose a novel approach 
dedicated to overcoming 
catastrophic forgetting problem in GNNs. 
Specifically, we introduce a generic module,
dubbed as topology-aware weight preserving~(TWP),
portable to arbitrary GNNs in a plug-and-play fashion. 
As opposed to conventional CNN-based methods that 
strive to retain only the performances of the old
task by decelerating changes of the crucial parameters,
TWP also explicitly looks into topological aggregation,
the core mechanism that yields the success of GNNs,
and endeavors to preserve the attention
and aggregation strategy of GNNs on the old task.
In other words, TWP expressly accounts for
not only the node-level learning 
in terms of parameter updates,
but also the brief propagation between nodes
for the continual learning process.

\begin{figure*}[t]  
\centering
\includegraphics[width=0.9\textwidth]{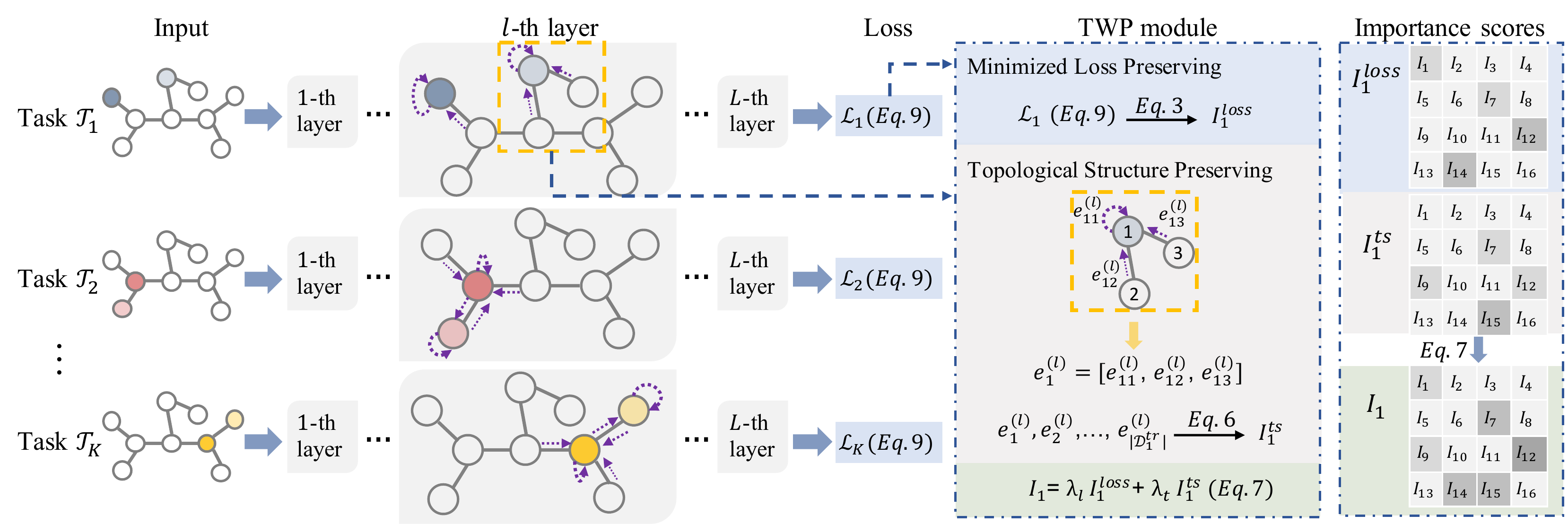}
\caption{Overview of the proposed method. 
For simplicity, we assume that there are only two classes in each task, 
and take task $\mathcal{T}_1$ as an example to illustrate the TWP module.
$I_1$ contains the importance scores of all network parameters for task $\mathcal{T}_1$, where different shades of color represent different important scores.
Best viewed in color.
}
\vspace{-1.0em}
\label{overview}
\end{figure*}

The workflow of the proposed GNN-based
continual learning approach is shown in Figure~\ref{overview}. 
Given an input graph and embedded feature of nodes, 
the TWP module estimates an importance score for each 
parameter of the network based on its contribution 
to both the task-related performance and the topological structure.
This is achieved by computing the 
gradients of the task-wise objective and 
the topological-preserving one
with respect to each parameter,
and treating such gradient as
an index for the parameter importance.
Upon learning a new task, we penalize
the changes to the significant parameters 
with respect to all the old tasks,
and hence enables us to ``remember'' the knowledge 
learned from the previous tasks.
{When employing our proposed method to
solve the problem in
Figure~\ref{cf}, the test accuracy of task~1 dropped by
only 3.77$\%$, which significantly enhances over the CNN-based method and 
showcases the power of utilizing the
topological information of graphs.}
Since TWP does not impose assumptions on the 
form of a GNN model,
it can be readily applicable to arbitrary GNN
architectures. 

We conduct extensive experiments to 
demonstrate the effectiveness
of the proposed model:
we evaluate the proposed method 
on three popular GNN architectures, 
namely {graph attention networks (GATs)~\cite{velivckovic2017graph}, graph convolutional networks (GCNs)~\cite{kipf2016semi}, and graph isomorphism networks (GINs)~\cite{xu2018powerful}}
over  four node classification datasets including 
Corafull, Amazon Computers, PPI, and Reddit, 
and one graph classification dataset, Tox21.
Results show that our method consistently achieves
the best performance among all the compared methods.

Our contribution is therefore 
introducing a novel
continual learning scheme tailored
for GNNs, dedicated to overcoming  
catastrophic forgetting.
The proposed topology-aware weight preserving (TWP) module
explicitly looks into the topological
aggregation mechanism of the learned tasks
to strength continual learning,
and serves as a portable block 
readily applicable to arbitrary GNN architectures. 
Experiments on three GNN backbones over five datasets
demonstrate that TWP consistently yields 
gratifying performances.

\section{Related work}

\subsubsection{Graph Neural Networks.}
A pioneering graph-based convolutional network approach was proposed in \cite{bruna2013spectral}, which generalized CNNs to signals defined on more general domains, and the work~\cite{defferrard2016convolutional} improved it 
by using Chebyshev polynomials. 
\cite{kipf2016semi} showed that GNNs can be built by stacking layers of first-order Chebyshev polynomial filters.
\cite{hamilton2017inductive}, on the other hand, proposed a framework based on sampling and aggregation instead of using all nodes,
while \cite{velivckovic2017graph} introduced an attention mechanism to GNNs 
by specifying different weights to different  neighboring nodes.
Many other GNN models~\cite{spagan,yang2020factorizable} 
have been explored recently, demonstrating promising performances
on graph applications~\cite{qiu2020hallucinating,WangECCV14}

\subsubsection{Continual Learning.}
This line of work has focused on the
memory modules~\cite{lopez2017gradient,rebuffi2017icarl,chaudhry2018efficient},
and thus falls within the domain of model reuse
~\cite{yang2020distilling,Ye_2019_CVPR,ShenAAAI19}.
However, the computation and memory costs increase rapidly as the number of tasks increases, which facilitated the emergence of pseudo-rehearsal methods~\cite{robins1995catastrophic} and  
generative network based methods~\cite{shin2017continual,nguyen2017variational}.
Recently, several methods have been proposed based on meta-learning algorithms~\cite{MER,javed2019meta} and online sample selection~\cite{aljundi2019gradient}.
Despite the great progress of experience replay methods,  
the dependency of nodes on a graph makes it difficult 
to selectively sample individual nodes as experience.

Another popular strategy for overcoming catastrophic forgetting focuses on preserving the parameters inferred in one task while training on another.
One milestone is Elastic Weight Consolidation (EWC)~\cite{kirkpatrick2017overcoming} 
which slows down learning on important parameters for  previous task.
Subsequently, many other regularization based methods have been proposed~\cite{zenke2017continual, lee2017overcoming, li2017learning, aljundi2018memory, mallya2018packnet} to help remember the old knowledge by penalizing changes to important parameters for previous tasks. 
The last major family seeks to prevent forgetting of  old tasks by assigning different parameter subsets for different tasks and reusing as much previous knowledge as possible~\cite{french1994dynamically, chen2015net2net, rusu2016progressive, schwarz2018progress,  serra2018overcoming}.  
Our method can be cast as a regularization-based method, 
as it attempts to preserve  parameters inferred in  
previous tasks to overcome catastrophic forgetting.
Unlike previous methods, however, we explicitly 
integrate  topological  information of graphs into 
continual learning.

\section{Preliminaries}
Before introducing the proposed approach, we briefly
review graph
neural networks 
(GNNs) and formulate the problem 
of continual learning on GNNs.

\subsubsection{Graph Neural Networks.}
Given a graph 
$\mathcal{G}=\{\mathcal{V},\mathcal{E}\}$ 
with $N$ nodes, specified as a set of node feature $X =\{\boldsymbol{x}_1, \boldsymbol{x}_2,...,\boldsymbol{x}_N\}$ 
and an adjacency matrix $\boldsymbol{A}$ 
that represents the adjacency relations among nodes. 
The hidden representation of node $v_i$ at the $l$-th layer, denoted as $\boldsymbol{h}_i^{(l)}$, is computed by:
\begin{equation}
\label{gcn}
  \boldsymbol{h}_i^{(l)}=\sigma(\sum_{j \in \mathcal{N}(i)}
   \mathcal{A}_{ij} \boldsymbol{h}_j^{(l-1)} W^{(l)}),
\end{equation}
\noindent where $\mathcal{N}(i)$ denotes the neighbors of node $v_i$,
$\sigma(\cdot)$ is an activation function,
and $W^{(l)}$ is the transformation matrix of the $l$-th layer.
$\boldsymbol{h}_i^{(0)}$ represents the input feature of node $v_i$.
$\mathcal{A}$ is a matrix that defines the aggregation strategy 
from neighbors,
which is one of the cores of GNNs.

The graph convolutional networks~(GCNs)~\cite{kipf2016semi} 
computes $\mathcal{A}$ based on 
a first-order approximate of the spectral of graph, which is fixed
given $\boldsymbol{A}$.
For attention-based GNNs, 
like graph attention networks~(GATs)~\cite{velivckovic2017graph}, 
$\mathcal{A}$ is computed based on the pair-wise attention,
which is defined as
\begin{equation}
\label{eij}
  e_{ij}^{(l)}=S_{j \in \mathcal{N}(i)} a(\boldsymbol{h}_i^{(l-1)} W^{(l)},\boldsymbol{h}_j^{(l-1)} W^{(l)}),
\end{equation}
\noindent where $a$ is a neural network, $S$ represents softmax normalization.
Authors in~\cite{xu2018powerful} further invest the expression power of GNNs
and proposes graph isomorphism networks~(GINs).

\subsubsection{Problem Formulation.}
In a learning sequence, the model
receives a sequence of disjoint tasks
$\mathcal{T}=\{\mathcal{T}_1,\mathcal{T}_2,...,\mathcal{T}_K\}$ which will be learned sequentially. 
Each task $\mathcal{T}_k$ contains a training node set $\mathcal{V}_k^{tr}$ and a testing node set $\mathcal{V}_k^{te}$, with corresponding feature sets $X_k^{tr}$ and $X_k^{te}$.
Each node $v_i \in \{\mathcal{V}_k^{tr} \cup \mathcal{V}_k^{te} \}$ corresponds a category label $y^l \in \mathcal{Y}_k$ where $\mathcal{Y}_k = \{y^1, y^2,...,y^{c_k} \}$ is the label set and $c_k$ is the number of classes in task $\mathcal{T}_k$. 
In the continual learning settings, different tasks correspond to different splits of a dataset without overlap in category labels.
Once the learning of a task is completed, the data related to this task is no longer available.
In this paper, we aim to learn a shared GNN model $f_W$ parameterized by $W=\{\boldsymbol{w}_m\}$ over a sequence of graph related tasks $\mathcal{T}$,  
such that this model can not only perform well on the new task but also remember old tasks.

\section{Method}
In this section, 
we provide the details of
the topology-aware weight preserving (TWP) module
and how the module boost
the continual learning performance of
GNNs.
Moreover, we impose sparse regularization to
balance the plasticity and stability of GNNs.
We also provide the scheme to extend the proposed 
method to arbitrary GNNs.

\subsection{Topology-aware Weight Preserving}
The TWP module captures the topology information of graphs 
and finds the crucial parameters that are both important 
to the task-related objective and the topology-related one. 
It contains two sub-modules: 
minimized loss preserving and topological structure preserving.

\subsubsection{Minimized Loss Preserving.} 
After training task $\mathcal{T}_k$, the model has learned 
a set of optimal parameters $W_k^*$ 
that minimize the loss of this task.
However, not all network parameters contribute equally.
We thus try to find the crucial parameters for 
preserving the minimized loss.

Given the training set $\mathcal{D}_k^{tr}$
for task $\mathcal{T}_k$, 
we denote the loss as $\mathcal{L}(X_k^{tr};W)$, 
where $X_k^{tr}$ contains the feature of nodes in $\mathcal{D}_k^{tr}$ and $W=\{\boldsymbol{w}_m\}$ contains all network parameters. 
Like~\cite{zenke2017continual}, the change in loss for an infinitesimal parameter perturbation $\Delta W =\{\Delta \boldsymbol{w}_m\}$ can be approximated by
\begin{equation}
\label{loss}
  \mathcal{L}(X_k^{tr};W+ \Delta W)-\mathcal{L}(X_k^{tr};W) \approx \sum_m f_m(X_k^{tr}) \Delta \boldsymbol{w}_m,
\end{equation}
where $\Delta \boldsymbol{w}_m$ is the infinitesimal change in parameter $\boldsymbol{w}_m$.  $f_m(X_k^{tr})=\frac{\partial \mathcal{L}}{\partial{\boldsymbol{w}_m}}$ is the gradient of the loss with respect to the parameter $\boldsymbol{w}_m$, and it can be used to approximate the contribution of the parameter $\boldsymbol{w}_m$ to the loss.

To remember the task $\mathcal{T}_k$ while learning future tasks, we expect to preserve the 
minimized loss as much as possible, 
which can be achieved by
maintaining the stability of parameters that are important for the minimized loss while learning future tasks.
Based on Equation~\ref{loss}, 
we represent the importance of a parameter $\boldsymbol{w}_m$ 
as the magnitude of the gradient $f_m$.  
The importance of all parameters $W$ to the loss
of task $\mathcal{T}_k$ can be denoted as 
$I_{k}^{loss}=[\left \|f_m(X_k^{tr})\right \|]$, 
where $I_{k}^{loss}$ is a matrix contains the importance scores of all parameters.

\subsubsection{Topological Structure Preserving.}
Since the topological information of a graph plays 
an important role in the aggregation strategies, 
we seek to find parameters that
are important to the learned topological information.
To give a better illustration,
we first adopt GATs as the base model and model
the attention coefficients between the center node and its neighbors as the topological information around the center node.
We rewrite the attention coefficient $e_{ij}^{(l)}$ between node $v_i$ and its neighbor $v_j$ at the $l$-th layer, as shown in Equation \ref{eij}, in the form of matrix:
\begin{equation}
\label{attmat}
  e_{ij}^{(l)} = a(\boldsymbol{H}_{i,j}^{(l-1)}; W^{(l)}),
\end{equation}
\noindent where $a$ is a neural network and $W^{(l)}$ is the weight matrix of the $l$-th layer.
$\boldsymbol{H}_{i,j}^{(l-1)}$ contains the embedding feature of node $v_i$ and $v_j$ computed from the $(l-1)$-th layer.
Similar to Equation \ref{loss}, the change in $e_{ij}^{(l)}$ caused by the infinitesimal change $\Delta W=\{ \Delta \boldsymbol{w}_m \}$ 
can be approximated by:
\begin{equation}
\begin{split}
\label{attsen}
  a(\boldsymbol{H}_{i,j}^{(l-1)}; W^{(l)} + \Delta W^{(l)})
  -a(\boldsymbol{H}_{i,j}^{(l-1)}; W^{(l)}) \\
  \approx \sum_m g_m (\boldsymbol{H}_{i,j}^{(l-1)}) \Delta \boldsymbol{w}_m,
\end{split}
\end{equation}
\noindent where $g_m (\boldsymbol{H}_{i,j}^{(l-1)})=\frac{\partial a}{\partial{\boldsymbol{w}_m}}$
is the gradient of the attention coefficient $e_{ij}^{(l)}$ 
with respect to the parameter $\boldsymbol{w}_m$.

Unlike the task-related loss that is already defined and is a single value,
there is no objective function for the attention layer and 
the attention coefficients centered on node $v_i$ at the $l$-th layer form a multi-dimensional vector $\boldsymbol{e}_i^{(l)} =[ e_{i1}^{(l)},..., e_{i |\mathcal{N}_i |}^{(l)}]$.  
We define the topological loss for all nodes in $\mathcal{D}_k^{tr}$ as the squared $l_2$ norm of the multi-dimensional vector $\boldsymbol{e}_i^{(l)}$ at the $l$-th layer
and compute the gradients as
\begin{equation}
\label{vi}
  g_m(\boldsymbol{H}^{(l-1)})=\frac{\partial \left ( \left| [e_1^{(l)},...,e_{|\mathcal{D}_k^{tr}|}^{(l)}] \right |_2^2 \right ) }{\partial{\boldsymbol{w}_m}},
\end{equation}
\noindent 
where $\boldsymbol{H}^{(l-1)}$ is the output of the $(l-1)$-th layer. 
The importance scores of all parameters 
to the topological structure in task $\mathcal{T}_k$ 
can be represented as 
$I_{k}^{ts}=[ \|g_m(\boldsymbol{H}_k^{(l-1)}) \|]$, 
where $I_{k}^{ts}$ is a matrix contains the importance scores of all parameters.
In the experiments, we always adopt the attention coefficients 
computed in the middle layer of the GNN model
as the topological information of graphs.

The final importance scores of all parameters $W$ to the task $\mathcal{T}_k$ can be computed by
\begin{equation}
\label{Ik}
    I_k =\lambda_l I_{k}^{loss} + \lambda_t I_k^{ts},
\end{equation}
\noindent where $\lambda_l$ and $\lambda_t$ are two hyper-parameters which together determine the importance of parameters.
Changing of parameters with small importance scores 
does not affect the performance of the task much, 
and can minimize the loss for subsequent tasks.

\subsection{Continual Learning on GNNs}
After learning each task, we can measure the 
importance of all the parameters 
by using the proposed TWP module. 
When learning a new task $\mathcal{T}_{k+1}$, 
in addition to improve the performance of
the new task, 
we also need to remember old ones 
by maintaining the stability of 
the important parameters for them, which can be achieved 
by penalizing changes to important parameters for old tasks.
The loss function for the new task $\mathcal{T}_{k+1}$ is
formulated as
\begin{equation}
\label{loss2}
  \mathcal{L}^{'}_{k+1}(W) = \mathcal{L}_{k+1}^{new}(W) +  \sum_{n=1}^{k} I_n \otimes (W - W^*_{n})^{2},
\end{equation}
\noindent where $\otimes$ means element-wise multiplication. 
$\mathcal{L}_{k+1}^{new}(W)$ is the 
task-related loss function, 
e.g., cross entropy 
loss for classification task.
$I_n$ indicates how important the network parameters 
for the old task, 
and $W_n^*$ contains the optimal parameters for 
the task $\mathcal{T}_n$.
Such learning strategy ensures that parameters with low importance scores 
are free to change to adapt to the new task, 
while changing the parameters with high importance scores is penalized
so that the model still performances well on  previous tasks.

\subsection{Promoting Minimized Importance Scores}
Although maintaining the stability of the important parameters can make the model
remember the learned tasks, there is a risk that the model
will have less plasticity and cannot learn well on the subsequent tasks.
In order to have sufficient model capacity reserved for the future tasks,
we promote minimized on the computed importance scores 
of all parameters by adding 
a $l_1$ norm as a regularization to the importance scores 
computed from the current task.
Then the total loss while training the current task 
$\mathcal{T}_{k+1}$ will become
\begin{equation}
\label{totalloss}
  \mathcal{L}_{k+1}(W) = \mathcal{L}^{'}_{k+1}(W) + \beta  \left\| I_{k+1}  \right\|_1,
\end{equation}
\noindent where $\beta$ is a hyper-parameter that 
controls the capacity preserved for future tasks. 
$I_{k+1}$ contains the importance scores of all parameters 
for current task, which is 
computed based on the proposed TWP module.
$\beta$ affects the preserved capacity for future tasks. 
Higher $\beta$ will preserve larger learning capacity for
future tasks.

\subsection{Extension to General GNNs}
\begin{figure}[t]  
\vspace{-1.0em}
\setlength{\abovecaptionskip}{0.05cm}
\centering
\includegraphics[width=0.35\textwidth]{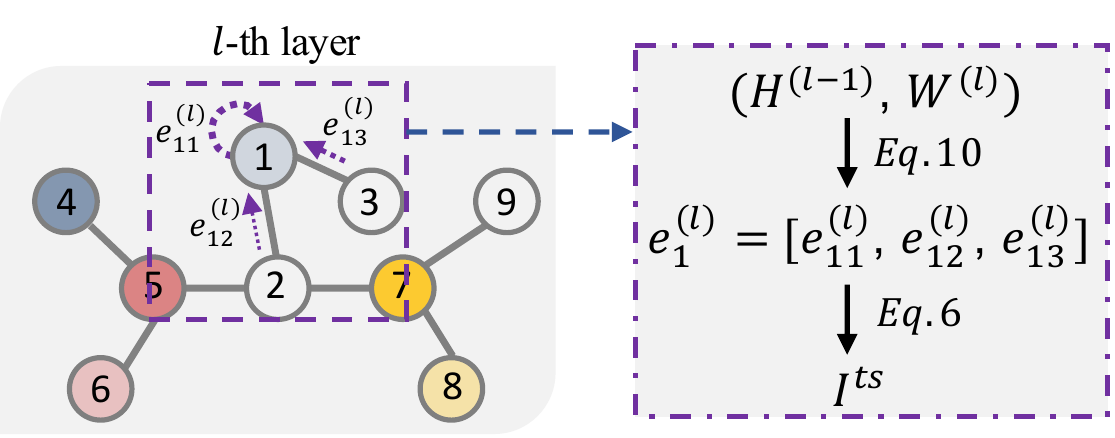}
\caption{Extension to general GNN models.}
\vspace{-1.0em}
\label{att}
\end{figure}

So far, we have discussed how to improve the continual learning performance
on the graph domain for the attention-based graph model.
We show in this section that the proposed method can
be easily extended to GNNs that may not hold an attention mechanism, 
such as GCNs and GINs,  
which makes it ready for arbitrary GNNs.
As shown in Figure~\ref{att}, we construct the topological structure
by adding a non-parametric attention mechanism, 
which is formulated as
\begin{equation}
\label{gcnatt}
  e_{ij}^{(l)} = (\boldsymbol{h}_i^{(l-1)} W^{(l)})^\mathrm{T}
  {\rm tanh}
  (\boldsymbol{h}_j^{(l-1)} W^{(l)}),
\end{equation}
This makes the attention weights dependent on the distance between the node $v_i$ and its neighbor $v_j$. 
Then, we normalize these distances 
across all neighbors of node $v_i$ as the final coefficients. 
Given the constructed topological structure, 
we can directly employ the proposed TWP module to
an arbitrary GNNs.
Notice that although we construct the attention coefficients between nodes, we only use them for the TWP module.
The models still use their original graph to aggregate 
and update the feature of nodes.

\begin{table*}[]
\setlength{\abovecaptionskip}{0.05cm}
\setlength{\belowcaptionskip}{-0.0cm}
\caption{Performance comparison with different GNN backbones on different datasets. 
For the task performance, we use classification accuracy on Corafull and Amazon Computers datasets, and micro-averaged F1 score for PPI and Reddit datasets. 
The symbol $\uparrow$~($\downarrow$) indicates higher~(lower) is better.
We highlight the \textcolor{red}{best} and \textcolor{blue}{second best}
in red and blue respectively.}
\centering
\scalebox{0.75}{%
\begin{tabular}{c|c|lc|cc|cc|cc}
\toprule
\multirow{2}{*}{\begin{tabular}[c]{@{}c@{}}Base\\ Models\end{tabular}} & \multirow{2}{*}{Methods} & \multicolumn{2}{c|}{Corafull}                                     & \multicolumn{2}{c|}{Amazon Computers}                                                      & \multicolumn{2}{c|}{PPI}                                                                  & \multicolumn{2}{c}{Reddit}                                        \\ \cline{3-10} 
                                                                       &                          & $\quad$ AP ($\uparrow$)                             & AF ($\downarrow$)                           & AP ($\uparrow$)                              & AF  ($\downarrow$)                          & AP ($\uparrow$)                             & AF ($\downarrow$)                           & AP ($\uparrow$)      & AF  ($\downarrow$)                          \\
\toprule
\multirow{7}{*}{GATs}                                                   & Fine-tune                 & 51.6 $\pm$ 6.4$\%$ & 46.1 $\pm$ 7.0$\%$                          & 86.5 $\pm$ 8.0$\%$                           & 12.3 $\pm$ 12.3$\%$                         & 0.365 $\pm$ 0.024                           & 0.178 $\pm$ 0.019                           & 0.474 $\pm$ 0.006    & 0.580 $\pm$ 0.007                           \\ 
                                                                       & LWF                      & 57.3 $\pm$ 2.3$\%$ & 39.5 $\pm$ 3.1$\%$                          & 90.3 $\pm$ 6.4$\%$                           & 9.9 $\pm$ 7.0$\%$                           & 0.382 $\pm$ 0.024                           & 0.185 $\pm$ 0.060                           & 0.500 $\pm$ 0.033    & 0.550 $\pm$ 0.034                           \\
                                                                       & GEM                      & 84.4 $\pm$ 1.1$\%$ & \textcolor{blue}{4.2} $\pm$ 1.0$\%$                           & \textcolor{blue}{97.1} $\pm$ 0.9$\%$                           & \textcolor{blue}{0.7} $\pm$ 0.5$\%$                           & 0.741 $\pm$ 0.016                           & 0.112 $\pm$ 0.030                           & \textcolor{blue}{0.947} $\pm$ 0.001    &  \textcolor{blue}{0.030} $\pm$ 0.008                          \\
                                                                       & EWC                      & \textcolor{blue}{86.9} $\pm$ 1.7$\%$ & 6.4 $\pm$ 1.8$\%$                           & 94.5 $\pm$ 3.3$\%$                           & 4.6 $\pm$ 4.5$\%$                           & \textcolor{blue}{0.826} $\pm$ 0.027                           & 0.142 $\pm$ 0.028                           & 0.944 $\pm$ 0.019    &0.032 $\pm$ 0.021                           \\
                                                                       & MAS                      & 84.1 $\pm$ 1.8$\%$ & 8.6 $\pm$ 2.2$\%$                           & 94.0 $\pm$ 5.5$\%$                           & 5.0 $\pm$ 6.9$\%$                           & 0.749 $\pm$ 0.007                           & \textcolor{blue}{0.092} $\pm$ 0.008                           & 0.865 $\pm$ 0.031    & 0.085 $\pm$ 0.024                           \\
                                                                       & Ours                     & \textcolor{red}{89.0} $\pm$ 0.8$\%$ & \textcolor{red}{3.3} $\pm$ 0.3$\%$ & \textcolor{red}{97.3} $\pm$ 0.6$\%$ & \textcolor{red}{0.6} $\pm$ 0.2$\%$ & \textcolor{red}{0.853} $\pm$ 0.004 & \textcolor{red}{0.086} $\pm$ 0.005 & \textcolor{red}{0.954} $\pm$ 0.014    & \textcolor{red}{0.014} $\pm$ 0.015 \\ 
                                                                       & Joint train              & 91.9 $\pm$ 0.8$\%$ & 0.1 $\pm$ 0.2$\%$                           & 98.2 $\pm$ 0.6$\%$                           & 0.02 $\pm$ 0.1$\%$                          & 0.931 $\pm$ 0.040                           & 0.035 $\pm$ 0.026                           & 0.978 $\pm$ 0.001    & 0.001 $\pm$ 0.001                           \\ \bottomrule
                                                                        \toprule
\multirow{7}{*}{GCNs}                                                   & Fine-tune                 & 34.0 $\pm$ 3.1$\%$ & 63.3 $\pm$ 2.8$\%$                          & 78.0 $\pm$ 6.3$\%$                           & 24.7 $\pm$ 6.8$\%$                         & 0.366 $\pm$ 0.039                           & 0.294 $\pm$ 0.010                           & 0.397 $\pm$ 0.064    & 0.670 $\pm$ 0.072                           \\ 
                                                                       & LWF                      & 43.1 $\pm$ 5.8$\%$ & 53.5 $\pm$ 7.2$\%$                          & 86.2 $\pm$ 5.6$\%$                           & 14.3 $\pm$ 5.4$\%$                           & 0.540 $\pm$ 0.026                           & 0.338 $\pm$ 0.026                           & 0.441 $\pm$ 0.100    & 0.615 $\pm$ 0.113                           \\
                                                                       & GEM                      & 80.5 $\pm$ 2.1$\%$ & 5.5 $\pm$ 2.4$\%$                           & \textcolor{blue}{97.1} $\pm$ 1.2$\%$                           & \textcolor{blue}{0.9}  $\pm$ 0.3$\%$                           & 0.628 $\pm$ 0.114                           & 0.052 $\pm$ 0.086                           & 0.970 $\pm$ 0.004    & 0.014 $\pm$ 0.005                            \\
                                                                                                           & EWC                      & 84.5 $\pm$ 3.0$\%$ & 7.1 $\pm$ 2.3$\%$                           & 94.4 $\pm$ 2.5$\%$                           & 4.0 $\pm$ 3.4$\%$                           & 0.653 $\pm$ 0.009                           & 0.050 $\pm$ 0.012                           & 0.917 $\pm$ 0.029    & 0.074 $\pm$ 0.007                           \\
                                                                                                           & MAS                      & \textcolor{blue}{86.1} $\pm$ 1.9$\%$ & \textcolor{blue}{4.8} $\pm$ 1.3$\%$                           & 96.8 $\pm$ 1.2$\%$                           & 1.3 $\pm$ 0.8$\%$                           & \textcolor{blue}{0.656} $\pm$ 0.006                           & \textcolor{red}{0.014} $\pm$ 0.010                           & \textcolor{blue}{0.975} $\pm$ 0.002    & \textcolor{blue}{0.002} $\pm$ 0.060                           \\
                                                                       & Ours                     & \textcolor{red}{87.8} $\pm$ 1.5$\%$ & \textcolor{red}{2.9} $\pm$ 1.1$\%$ & \textcolor{red}{97.5} $\pm$ 0.6$\%$ & \textcolor{red}{0.6} $\pm$ 0.9$\%$ & \textcolor{red}{0.661} $\pm$ 0.010 & \textcolor{blue}{0.038} $\pm$ 0.012 & \textcolor{red}{0.976} $\pm$ 0.002    & \textcolor{red}{0.001} $\pm$ 0.062 \\ 
                                                                       & Joint train              & 88.6 $\pm$ 1.5$\%$ & 1.6 $\pm$ 1.0$\%$                           & 98.3 $\pm$ 0.4$\%$                           & 0.2 $\pm$ 0.3$\%$                          & 0.768 $\pm$ 0.003                           & 0.003 $\pm$ 0.006                           & 0.978 $\pm$ 0.003    & 0.001 $\pm$ 0.002                           \\ \bottomrule
                                                                        \toprule
\multirow{7}{*}{GINs}                                                   & Fine-tune                 & 30.9 $\pm$ 4.2$\%$ & 46.1 $\pm$ 7.0$\%$                          & 66.2 $\pm$ 7.1$\%$                           & 34.9 $\pm$ 8.6$\%$                         & 0.598 $\pm$ 0.033                           & 0.302 $\pm$ 0.002                           & 0.277 $\pm$ 0.068    & 0.382 $\pm$ 0.058                           \\ 
                                                                       & LWF                      & 34.1 $\pm$ 1.8$\%$ & 39.5 $\pm$ 3.1$\%$                          & 71.3 $\pm$ 3.3$\%$                           & 31.4 $\pm$ 7.8$\%$                           & 0.606 $\pm$ 0.011                           & 0.332 $\pm$ 0.014                           & 0.313 $\pm$ 0.059    & 0.250 $\pm$ 0.119                           \\
                                                                       & GEM                      & \textcolor{blue}{78.0} $\pm$ 2.8$\%$ & \textcolor{blue}{3.5} $\pm$ 0.6$\%$                           & 91.9 $\pm$ 5.0$\%$                           & 4.2 $\pm$ 2.7$\%$                           & \textcolor{blue}{0.710} $\pm$ 0.073                           & 0.132 $\pm$ 0.002                           & \textcolor{blue}{0.443} $\pm$ 0.045    & \textcolor{blue}{0.135} $\pm$ 0.030                             \\
                                                                                                    & EWC                      & 68.9 $\pm$ 11.7$\%$ & 6.4 $\pm$ 1.8$\%$                           & 83.7 $\pm$ 6.2$\%$                           & 10.2 $\pm$ 6.1$\%$                           & 0.644 $\pm$ 0.144                           & \textcolor{blue}{0.038} $\pm$ 0.008                           & 0.394 $\pm$ 0.058    & 0.168 $\pm$ 0.132                          \\
                                                                                                    & MAS                      & 73.7 $\pm$ 4.3$\%$ & 8.6 $\pm$ 2.2$\%$                           & \textcolor{blue}{92.4} $\pm$ 3.2$\%$                           & \textcolor{red}{1.7} $\pm$ 3.4$\%$                           & 0.686 $\pm$ 0.008                           & 0.082 $\pm$ 0.008                           & 0.333 $\pm$ 0.063    & 0.184 $\pm$ 0.097                          \\
                                          
                                                                       & Ours                     & \textcolor{red}{78.4} $\pm$ 1.6$\%$ & \textcolor{red}{3.3} $\pm$ 0.3$\%$ & \textcolor{red}{92.6} $\pm$ 1.0$\%$ & \textcolor{blue}{3.8} $\pm$ 1.1$\%$ & \textcolor{red}{0.715} $\pm$ 0.005 & \textcolor{red}{0.013} $\pm$ 0.001 & \textcolor{red}{0.451} $\pm$ 0.043    & \textcolor{red}{0.130} $\pm$ 0.036 \\ 
                                                                       & Joint train              & 83.5 $\pm$ 1.6$\%$ & 0.1 $\pm$ 0.2$\%$                           & 95.5 $\pm$ 0.6$\%$                           & 0.7 $\pm$ 1.3$\%$                          & 0.809 $\pm$ 0.037                           & 0.010 $\pm$ 0.002                         & 0.582 $\pm$ 0.037    & 0.012 $\pm$ 0.028                           \\ \bottomrule
\end{tabular}%
}
\label{result}
\vspace{-1.0em}
\end{table*}

\section{Experiments}
We have performed comparative evaluation of our method against a wide variety of strong baselines, on four node classification datasets (transductive as well as inductive) and one graph classification dataset,
on which our method achieves 
{consistent better or on-par} performances.
In the following sections, we provide the details of datasets, baselines, experimental setup, the quantitative results and analysis.

\subsection{Datasets}
To provide a thorough evaluation, 
we adopt datasets related to both
node-level and graph-level tasks.
More information of the datasets and settings of graph continual learning 
can be found in the supplementary material.

\subsubsection{Node Classification.}
For transductive learning, we utilize two widely
used datasets named Corafull \cite{bojchevski2017deep} and Amazon Computers~\cite{mcauley2015image}.
We construct nine tasks with five classes per task
on the Corafull dataset. 
Each task is a five-way node classification task.
We conduct five tasks on Amazon Computers 
dataset and 
each task has two classes.
For inductive learning, we use two datasets: a protein-protein interaction (PPI) dataset~\cite{zitnik2017predicting} 
and a large graph dataset of 
Reddit posts~\cite{hamilton2017inductive}.
We follow the same dataset splitting protocol as~\cite{hamilton2017inductive}.
We perform 12 tasks for PPI and each task has ten classes;
we generate eight tasks on Reddit 
and each one has five classes.

\subsubsection{Graph Classification.}
We conduct experiment on one graph classification dataset 
named Tox21~\cite{huang2014profiling}, 
which contains qualitative toxicity measurements 
for 8014 compounds on 12 different targets. 
Each target results in a binary label.
Each target in this dataset will be used for one task,
leading to 12 tasks and each one is a binary classification task. 


\subsection{Baselines}
Since there is no continual learning method designed for GNN models, we implement several continual learning methods designed for CNN models and apply them on graph domain.
These methods include Fine-tune~\cite{girshick2014rich}, 
LWF~\cite{li2017learning}, EWC~\cite{kirkpatrick2017overcoming}, 
MAS~\cite{aljundi2018memory}, and GEM~\cite{lopez2017gradient}.
Among these methods, Fine-tune can be seen as the lower bound
without any continual learning mechanism.
We also add Joint train method~\cite{caruana1997multitask} 
as the approximate \textit{upper bound} of
continual learning since this method allows the access of data from all learned tasks
during training.
LWF utilizes the idea of knowledge distillation to remember the old tasks.
EWC and MAS remember previous tasks by constraining changes to the important parameters.
GEM alleviates forgetting based on an experience replay buffer 
which is used to 
avoid the increasing of losses associated to old tasks.

\subsection{Metrics}
The performance of graph continual learning can be evaluated from
two different aspects named average performance~(AP) 
and average forgetting~(AF)~\cite{lopez2017gradient}, 
corresponding to the {plasticity and stability} of the model. 
More information about AP and AF could be found in the supplementary material.
Basically, AP measures the average test performance across all learned tasks 
while AF measures the performance difference between after 
learning the particular task and after learning subsequent tasks. 
For the task performance,
we use accuracy on Corafull and Amazon Computers datasets,
micro F1 score on PPI and Reddit datasets, and 
AUC score on Tox21 dataset.

\subsection{Experimental Setup}
To validate the generalization ability,
we conduct experiments based on three GNN backbones:
GATs~\cite{velivckovic2017graph}, 
GCNs~\cite{kipf2016semi}, and 
GINs~\cite{xu2018powerful}.
Since GATs uses attention mechanism to weight the neighbor for each node, 
we can directly employ the TWP module on it.
For GCNs, we first compute the attention coefficients between each node 
and its neighbors according to Equation~\ref{gcnatt} and
then use the TWP module to estimate the important scores of parameters.
Considering GINs was proposed for the graph-level classification task, we first take the embedding feature before the pooling operation and feed them into a linear layer whose output acts as the 
prediction for the whole graph.
Then we can compute the importance of parameters in the same way as GCNs.
The detailed architecture of these three GNN backbones
can be found in the supplementary material.

Adam SGD optimizer is used and the initial learning rate 
is set to 0.005 for all the datasets.
The training epochs are set to 200, 200, 400, 30, and 100 for Corafull, Amazon Computers, PPI, Reddit, and Tox21 respectively.
Early stopping is adopted for PPI and Tox21 where
the patience is 10 for both of them.
The regularizer hyper-parameter for EWC and MAS is always set to 10,000. 
The episodic memory for GEM contains all training nodes for Corafull and PPI datasets, and 100, 1,000, 100 training nodes for Amazon Computers, Reddit, and Tox21 datasets, respectively.
For our method, $\lambda_l$ is always set to 10,000, $\lambda_t$ is selected from 100 and 10,000 for different datasets, and $\beta$ is selected from 0.1 and 0.01. 
We set $\beta$ globally for all tasks.
We run five time with random seed and report the mean
and standard deviation for all methods and datasets.

\subsection{Node Classification Task}
Table~\ref{result} shows the results of performance comparison with baselines in terms of AP and AF for node classification task.
We can clearly observe catastrophic forgetting in GNNs from the higher AF for Fine-tune. 
Our proposed method achieves best or second best performance across all datasets and GNN backbones.
It is worth noting that, in some cases (e.g., PPI), some methods (e.g., MAS) perform better than other methods (e.g., EWC) in terms of AF, but become worse regards AP.
These methods scarify much the performance 
of the new tasks in order to remember the old ones.

\begin{figure*}[t]
\centering
\includegraphics[width=1.0\textwidth]{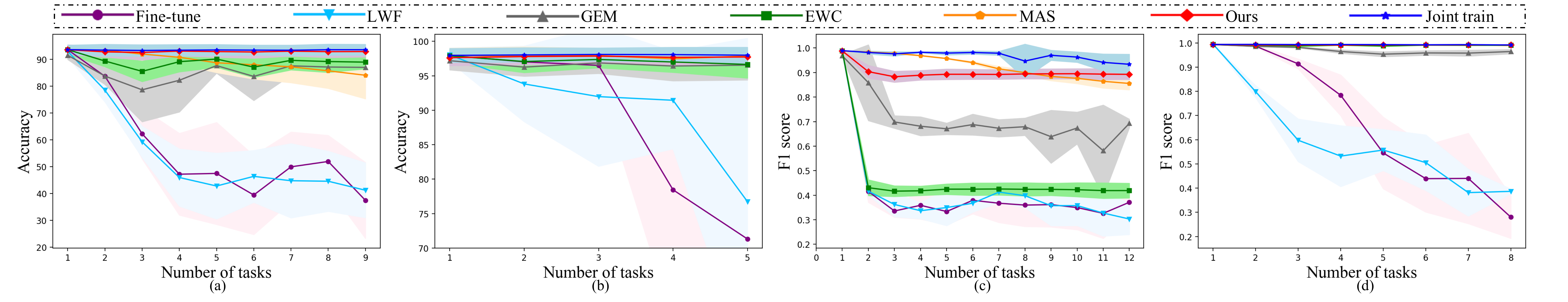}
\vspace{-2.2em}
\caption{Performance of the first task on Corafull~(a), 
Amazon Computers~(b), PPI~(c) and Reddit~(d), as more tasks are learned.}
\label{first}
\vspace{-0.6em}
\end{figure*}

\begin{figure*}[t]
\centering
\includegraphics[width=1.0\textwidth]{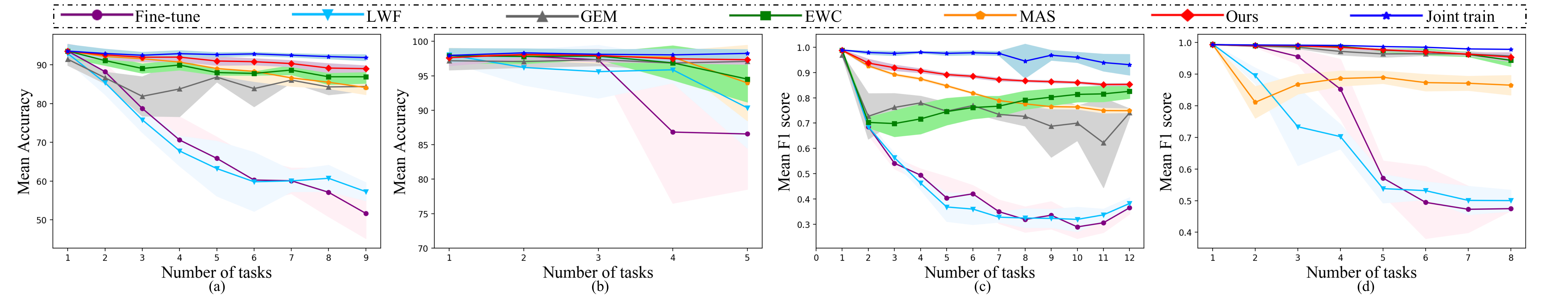}
\vspace{-2.2em}
\caption{Average performance across all the learned tasks on Corafull~(a), 
Amazon Computers~(b), PPI~(c), and Reddit~(d) as more tasks are learned.}
\label{mean}
\vspace{-1.6em}
\end{figure*}

Figure~\ref{first} shows the evolution of performance for the first task as more tasks are learned.
Figure~\ref{mean} shows the average performance across all the learned tasks
as more tasks are learned.
Our method performs minimal forgetting among all the methods, 
and is very close to upper-bound.
LWF, which adds new nodes to the network layer for each new task, performs inferior to other baselines even similar to the lower-bound in all the node classification datasets,
indicating that LWF is not adapted well to GNNs.
GEM, which prevents the loss for the previous tasks from increasing by storing a subset of the data of old tasks,
holds unstable performances across different datasets and GNN backbones. 
Although GEM achieves comparable results on Amazon Computers and Reddit dataset, it performs inferior to other baselines on Corafull and PPI even though we put all training nodes of each task into the episodic memory for the GATs backbone.
This can be partially explained by the fact that 
the stored data cannot play fully the role of remembering old knowledge.  
It is also worth noting that 
EWC and MAS, which aim to protect the important parameters 
when training the new task, 
work well on all these four datasets, which means that slowing down learning on important parameters is indeed applicable 
to GNNs.
Our method further boosts the performance over EWC and MAS, which demonstrates the superior of
the proposed TWP module in term of 
overcoming catastrophic forgetting on graph domain. 

\begin{figure}[t]
\setlength{\abovecaptionskip}{0.05cm}
\setlength{\belowcaptionskip}{-0.3cm}
\centering
\includegraphics[width=0.5\textwidth]{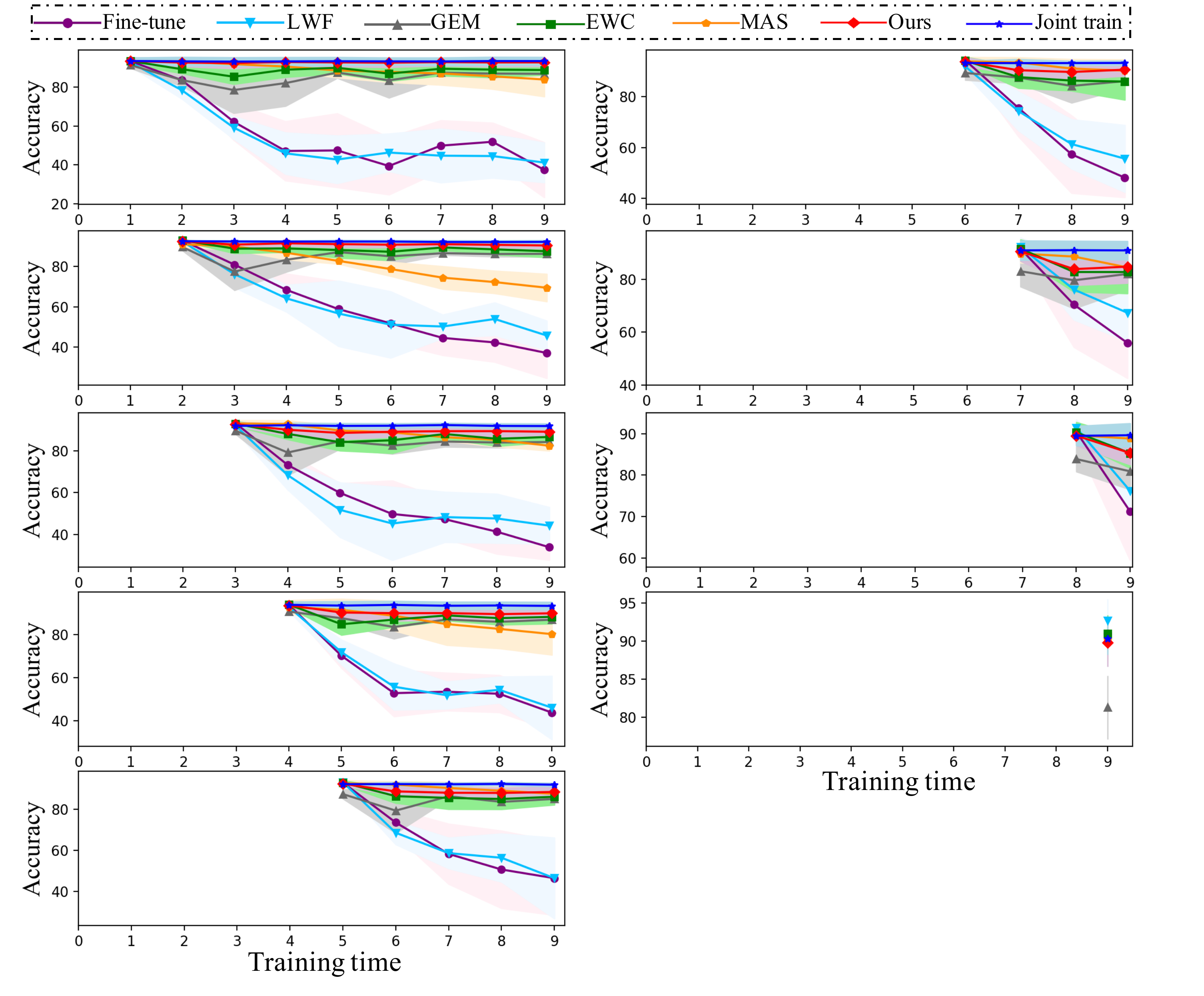}
\caption{
Evolution of performance for each task on GATs when gradually adding new tasks to the model. 
Different tasks are shown in different sub-graphs.
The $x$-axis denotes the new task added each time
and the $y$-axis denotes the performance of each task.
}
\vspace{-1.0em}
\label{traincora}
\end{figure}


\begin{figure}[!h]
\centering
\includegraphics[width=0.48\textwidth]{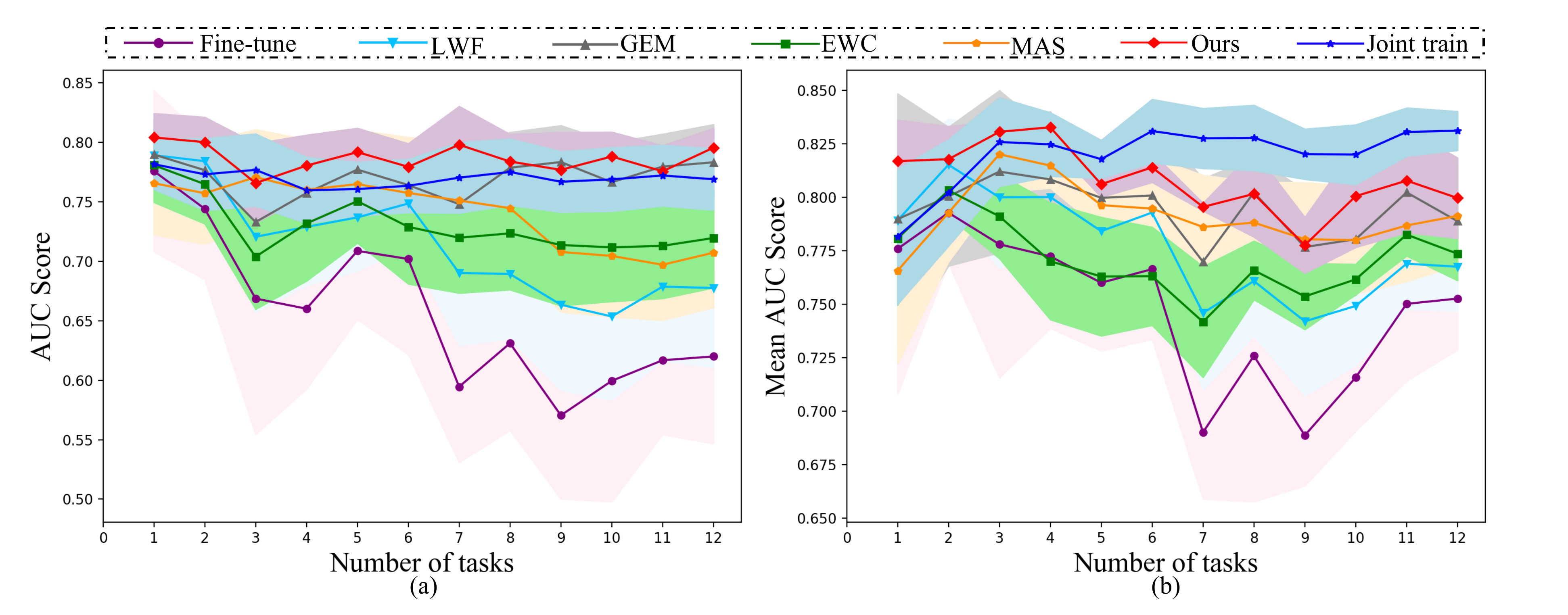}
\vspace{-2.2em}
\caption{Performance of the first task~(a) 
and average performance across all the learned tasks~(b)
as more tasks are learned
on Tox21 dataset based on GATs.
}
\vspace{-1.2em}
\label{tox}
\end{figure}

Figure~\ref{traincora} shows the training curves for nine tasks 
on Corafull dataset with GATs as the base model.
The performance of our method degrees slower over time than 
the baselines throughout the training process, 
and is generally similar to that of the joint training.
This is thanks to the capability of 
the proposed method to maintain
the learned topological information of previous tasks.

\subsection{Graph Classification Task}
The experiment of the graph-level task is shown in 
Table~\ref{auc_tox}, where Tox21 dataset is adopted. 
We use GATs and GCNs as the base models and
AUC as the metric to compute the continual learning
performance.
We also compare the performance of the first task and
the average performance across all the learned tasks
with GATs as the base model, which are
shown in (a) and (b) in Figure~\ref{tox} respectively.
Our method boosts the forgetting performance
by a significant margin, which demonstrates its effectiveness.

\begin{table}[!h]
\vspace{-0.3em}
\caption{Performance comparison on the Tox21 dataset.
}
\vspace{-0.4em}
\centering
\scalebox{0.66}{
\begin{tabular}{c|cc|cc}
\toprule
\multirow{2}{*}{Methods} & \multicolumn{2}{c|}{GATs} & \multicolumn{2}{c}{GCNs} \\ \cline{2-5} 
                         & AP ($\uparrow$)          & AF ($\downarrow$)        & AP ($\uparrow$)         & AF ($\downarrow$)        \\ \midrule
Fine-tune    & 0.753 $\pm$ 0.024 & 0.084 $\pm$ 0.021  & 0.737 $\pm$ 0.028 & 0.098 $\pm$ 0.011  \\
LWF         & 0.768 $\pm$ 0.041 & 0.058 $\pm$ 0.043  & 0.769 $\pm$ 0.034 & 0.059 $\pm$ 0.042  \\
GEM         & 0.789 $\pm$ 0.010 & 0.036 $\pm$ 0.017  & 0.768 $\pm$ 0.041 & 0.037 $\pm$ 0.006  \\
EWC         & 0.774 $\pm$ 0.025 & 0.038 $\pm$ 0.027  & 0.803 $\pm$ 0.031 & \textcolor{blue}{0.014} $\pm$ 0.015  \\
MAS         & \textcolor{blue}{0.790} $\pm$ 0.023 & \textcolor{blue}{0.032} $\pm$ 0.024  & \textcolor{blue}{0.805} $\pm$ 0.021 & \textcolor{blue}{0.014} $\pm$ 0.005  \\
Ours        & \textcolor{red}{0.801} $\pm$ 0.019 & \textcolor{red}{0.022} $\pm$ 0.012  & \textcolor{red}{0.810} $\pm$ 0.024 & \textcolor{red}{0.013} $\pm$ 0.012  \\
Joint train & 0.831 $\pm$ 0.009 & 0.006 $\pm$ 0.009  & 0.822 $\pm$ 0.017 & 0.012 $\pm$ 0.008  \\ \bottomrule
\end{tabular}}
\vspace{-1.0em}
\label{auc_tox}
\end{table}

\subsection{Ablation Study}
We conduct ablation study to validate the effectiveness of
each part of the proposed continual learning method.
First, we conduct experiment based purely on the original loss
of the task, 
namely, W/$\_$Loss in Table~\ref{ablation}.
In this case, our method would degrade into EWC, which measures the importance of parameters only based on the task related loss function.
Next, we verify the effectiveness of the topological information 
of graphs in graph-based continual learning 
by conducting experiments with the proposed TWP module.
According to Table~\ref{ablation}, TWP achieves consistent better 
performance on these two datasets, showing the effectiveness of preserving topological structure information to remember old tasks.
Finally, we promote minimized importance scores for all parameters to preserve model capacity for future tasks, 
which further improves overall performance. 

\begin{table}[!h]
\vspace{-0.4em}
\caption{Ablation study with GATs as the base model. 
W/$\_$Loss and W/$\_$TWP measures importance of parameters 
based on task-related loss and the proposed TWP module
respectively. Full is the proposed method with all parts.
}
\centering
\vspace{-0.4em}
\label{ablation}
\scalebox{0.76}{
\begin{tabular}{c|cc|cc}
\toprule
\multirow{2}{*}{Methods} & \multicolumn{2}{c|}{Corafull} & \multicolumn{2}{c}{Amazon Computers} \\ \cline{2-5} 
                         & AP ($\uparrow$)           & AF ($\downarrow$)           & AP ($\uparrow$)               & AF ($\downarrow$)                \\ \midrule
W/\_Loss                     & 86.9 $\pm$ 1.7$\%$   & 6.4 $\pm$ 1.8$\%$   & 94.5 $\pm$ 3.3$\%$   & 4.6 $\pm$ 4.5$\%$                   \\
W/\_TWP                      & 88.6 $\pm$ 1.0$\%$   & 3.7 $\pm$ 2.0$\%$   & 96.2 $\pm$ 2.2$\%$   & 0.8 $\pm$ 1.1$\%$                   \\
Full                     & 89.0 $\pm$ 0.8$\%$   & 3.3 $\pm$ 0.3$\%$  &  97.3 $\pm$ 0.6$\%$  & 0.6 $\pm$ 0.2$\%$                 \\ \bottomrule
\end{tabular}}
\vspace{-0.4em}
\end{table}

\section{Conclusion}
In this paper, we propose a dedicated continual learning method for graph neural networks, which is to our best knowledge the first attempt along this line.  
Specifically, we design a topology-aware weight preserving module which explicitly captures the topological information of graphs and measures the importance of the network's parameters based on the task-related loss function and the topological information.
When learning a new task, changes to the important parameters will be penalized to remember old tasks.
Moreover, the proposed approach can be 
readily extended to arbitrary GNNs.
The extensive experiments on both node-level tasks
and graph-level one 
demonstrates the effectiveness and applicability of 
the proposed continual learning method 
on the graph domain.

\newpage
\bibliography{aaai.bib}

\end{document}